\documentclass[10pt,twocolumn,letterpaper]{article}

\usepackage[pagenumbers]{iccv} 
\usepackage{times}
\usepackage{epsfig}
\usepackage{graphicx}
\usepackage{amsmath}
\usepackage{amssymb}
\usepackage{setspace}
\usepackage{makecell}
\usepackage{setspace}
\usepackage{mathcomp}
\usepackage{subcaption}
\usepackage{here}
\usepackage{multirow}
\usepackage{xcolor,colortbl}
\usepackage{algorithm}
\usepackage{algpseudocode}
\usepackage{tikz}
\usetikzlibrary{spy,backgrounds}
\usepackage{soul}

\usepackage[percent]{overpic}
\definecolor{iccvblue}{rgb}{0.21,0.49,0.74}
\usepackage[pagebackref,breaklinks,colorlinks,allcolors=iccvblue]{hyperref}

\def\paperID{*****} 
\def\confName{ICCV}
\def\confYear{2025}

\title{Objectness Similarity: Capturing Object-Level Fidelity in 3D Scene Evaluation}

\author{Yuiko Uchida, 
Ren Togo, 
Keisuke Maeda, 
Takahiro Ogawa,  
Miki Haseyama\\
Hokkaido University\\
{\tt\small \{uchida, togo, maeda, ogawa, mhaseyama\}@lmd.ist.hokudai.ac.jp}
}   

\begin{document}
\maketitle
\begin{abstract}
\label{sec:abstract}
This paper presents Objectness SIMilarity (OSIM), a novel evaluation metric for 3D scenes that explicitly focuses on ``objects," which are fundamental units of human visual perception. 
Existing metrics assess overall image quality, leading to discrepancies with human perception.
Inspired by neuropsychological insights, we hypothesize that human recognition of 3D scenes fundamentally involves attention to individual objects.
OSIM enables object-centric evaluations by leveraging an object detection model and its feature representations to quantify the ``objectness" of each object in the scene. 
Our user study demonstrates that OSIM aligns more closely with human perception compared to existing metrics. 
We also analyze the characteristics of OSIM using various approaches.
Moreover, we re-evaluate recent 3D reconstruction and generation models under a standardized experimental setup to clarify advancements in this field.
The code is available at \url{https://github.com/Objectness-Similarity/OSIM}. 
\end{abstract}    
\section{Introduction}
\label{sec:intro}

Recent advances in 3D reconstruction and generation have pushed the boundaries of visual quality, training efficiency, and rendering speed~\cite{nerf, mipnerf360, INGP, Zip-NeRF, 3dgs, dreamgaussian, one-2-3-45++, triplanegaussian, lgm}.
These breakthroughs open up a wide range of applications, including avatars, animation, extended reality, simulation, and robotics~\cite{nerf-survey, 3DGSsurvey, 3DGSsurvey2, 3DGSsurvey3}.
Despite these rapid advances, existing evaluation metrics remain insufficient to fully capture the progress.
Traditional metrics such as Peak Signal-to-Noise Ratio (PSNR) and Structural Similarity Index (SSIM)~\cite{ssim} are still the most widely used benchmarks. 
Learned Perceptual Image Patch Similarity (LPIPS)~\cite{lpips} captures high-level semantic fidelity using deep feature embeddings, whereas geometric measures such as Chamfer Distance (CD)~\cite{chamferdistance} assess the accuracy of generated shapes. 
Each metric has distinct advantages and disadvantages and is often used complementarily. 
Nevertheless, conventional metrics still struggle to accurately capture human subjective quality in 3D scene evaluation~\cite{evaluatingNVS, exact-nerf}.

\begin{figure}[t]
    \centering
    \includegraphics[width=\linewidth]{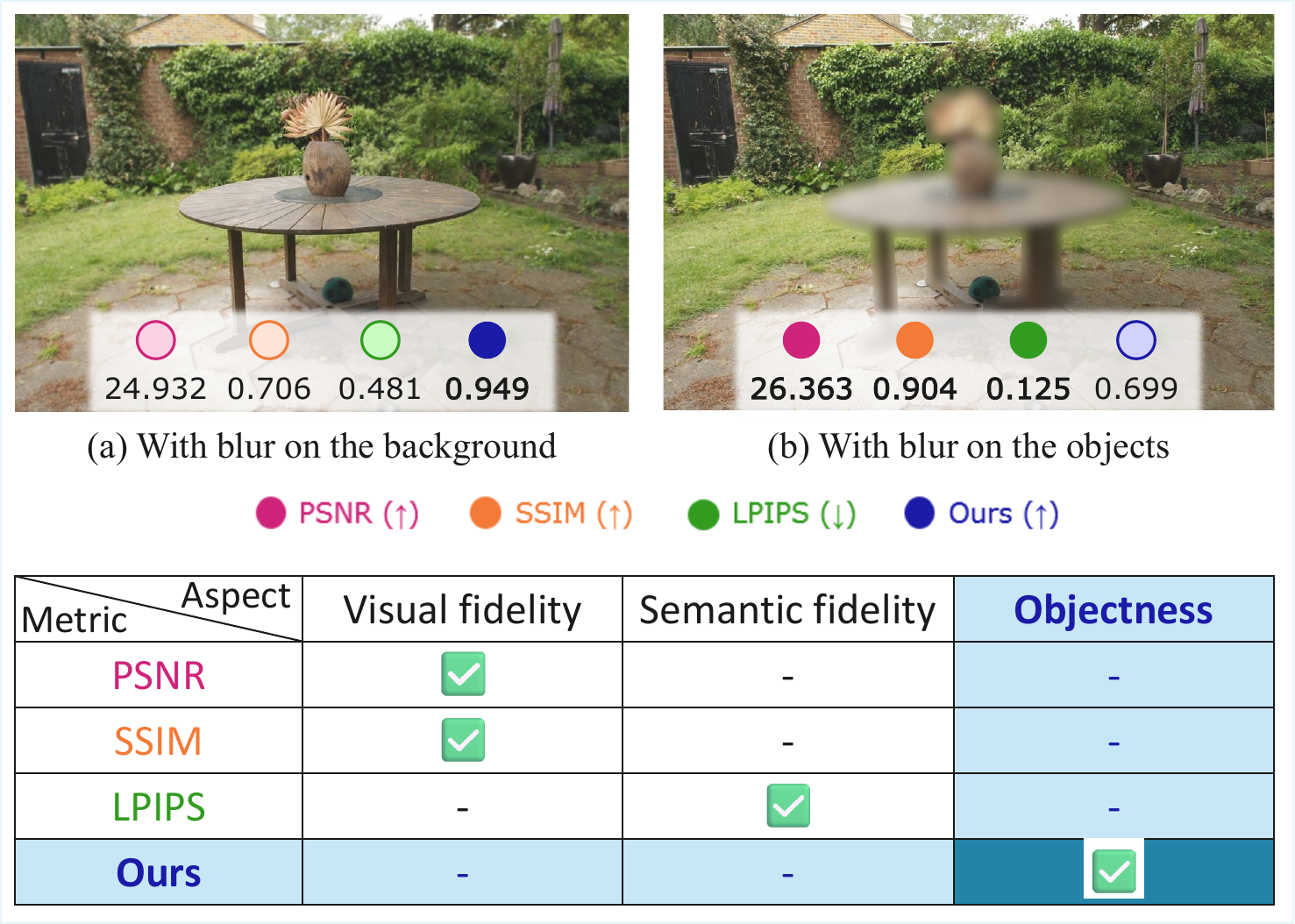}
    \caption{\small{(Top): An example of evaluation results. Darker dots indicate better scores. (Bottom): The evaluation perspective of each metric. Our proposed metric provides another evaluation perspective of object-centric evaluation. }}
    \label{fig:intro_fig}
\end{figure}

The top of Fig.~\ref{fig:intro_fig} illustrates the discrepancy between conventional evaluation metrics and human perception. 
The rendered image (a) has Gaussian blur added to the background, and the rendered image (b) has Gaussian blur added to the objects. 
According to human visual perception, image (a) is generally perceived as having higher quality than image (b). 
However, traditional metrics rate image (b) as higher quality, since they assess errors across the entire image rather than focusing on individual objects. 
Given that 3D reconstruction and generation results are ultimately judged by humans, evaluation metrics must closely align with human visual perception.
This limitation motivates us to introduce an additional evaluation metric specifically designed to assess individual objects within 3D scenes, thereby complementing existing metrics.

The significance of object-centric evaluation is further supported by neuropsychological research.
Studies indicate that human attention is inherently object-based: the brain first segments a scene into coherent, bounded entities and then preferentially selects, tracks, and encodes these objects as the fundamental units of perception~\cite{objects_and_attention}.
In daily life, humans instinctively focus on individual objects, recognizing each for its meaning and importance. 
Thus, accurately evaluating 3D scenes requires confirming that each object has sufficient quality to be identifiable as a specific entity. 
We refer to this criterion as ``objectness".
Object-focused evaluation is also important for machines interacting with 3D environments. 
For example, in robotic simulations, if a vision system cannot accurately recognize the object that a robotic arm must grasp, the task cannot be successfully executed.
As 3D scene applications continue to expand, object-level evaluation is becoming increasingly important.

In this paper, we propose \textit{Objectness SIMilarity (OSIM)}, a novel evaluation metric that explicitly introduces an object-centric perspective for 3D scene evaluation. 
OSIM leverages features from existing object detection models to evaluate each object's ``objectness,” thereby aligning evaluations more closely with human perception.
Additionally, by incorporating a saliency map, OSIM emphasizes the quality of objects naturally attracting human attention. 
Through comprehensive analysis and user studies, we demonstrate that OSIM provides evaluations more closely aligned with human perception.
As shown in the bottom of Fig.~\ref{fig:intro_fig}, OSIM effectively addresses the limitations of conventional metrics, enhancing overall 3D scene evaluation.  

Another contribution is our comprehensive re-evaluation of modern 3D methods under unified conditions. 
Although many improved models have been proposed, previous studies use different datasets, image resolutions, and metric parameters, making their reported gains hard to compare.
Building on the NerfBaselines benchmark~\cite{nerfbaselines}, we augment the reconstruction evaluation with mean opinion scores (MOS) and our proposed object-centric metric, then apply the same standardized protocol to leading 3D generation models. 
By addressing inconsistencies in datasets and evaluation protocols, our study enables a more consistent and reliable assessment of recent advances in this field.

In summary, our contributions are as follows:
\begin{itemize}
    \item We propose \textit{Objectness SIMilarity (OSIM)}, an object-centric metric that introduces a novel perspective to 3D scene evaluation. OSIM emphasizes the ``objectness" that humans inherently focus on. 
    \item Our user study demonstrates that incorporating the object-centric perspective of OSIM enables assessments that align more  closely with human perception. 
    \item We re-evaluate recent 3D reconstruction and generation models in a unified setting with various metrics, shedding light on the progress in this field.
\end{itemize}

\section{Related Work}
\label{sec:related_work}

Image-based 3D reconstruction and generation models have achieved remarkable progress in recent years, and attracted attention due to their wide-ranging potential applications, including animations, extended reality, simulation, and robotics~\cite{3DGSsurvey, 3DGSsurvey2, 3DGSsurvey3, nerf-survey, generation_survey}. 
These models reconstruct or generate photorealistic 3D representations from input images and are often used for Novel-View Synthesis (NVS) tasks that render images from angles not originally captured.  

Evaluation of these models relies on various metrics designed to assess different aspects.
Conventional metrics can be categorized into two types: image-based metrics and 3D shape-based metrics.
Image-based metrics include PSNR, SSIM~\cite{ssim}, LPIPS~\cite{lpips}, CLIP-similarity (CLIP-sim.)~\cite{clip} and Fr\'{e}chet Inception Distance (FID)~\cite{fid}. 
PSNR measures pixel-level fidelity by quantifying noise levels.
While simple and widely applicable, PSNR is highly sensitive to pixel-level errors and often correlates poorly with human perception~\cite{ssim, sheikh2006statistical}.
In contrast, SSIM and its multi-scale extension, MS-SSIM~\cite{ms-ssim} evaluate structural attributes such as contrast and texture.
Since structural information is crucial for the human visual system, SSIM aligns more closely with human perception.
However, SSIM still fails to capture higher-level semantic features~\cite{lpips}.
Deep learning–based metrics address this by incorporating semantic information.
LPIPS leverages features from pre-trained networks such as AlexNet~\cite{alexnet} and VGG~\cite{vgg}, which have been shown to correlate well with human perceptual judgments.  
CLIP-sim. quantifies semantic similarity by comparing image embeddings within a joint image–text feature space~\cite{clip}.
FID measures the similarity between the feature distributions of the generated and reference images using the Inception-V3 model~\cite{inceptionv3}. 
However, metrics based on pre-trained networks may inherit biases from their training data.
While existing metrics primarily focus on global image luminance or feature-based embeddings, our proposed metric complements them by incorporating an object-level evaluation. 
This novel perspective enables a more holistic and perceptually aligned assessment of 3D scenes.

3D shape-based metrics include Chamfer Distance (CD)~\cite{chamferdistance}, F-score~\cite{tanks_and_temples, fscore2}, and Volume IoU. 
These metrics evaluate how closely the generated 3D shape aligns with the ground truth by comparing predicted and original 3D models. Specifically, CD computes the average shortest distance between points of two models, quantifying shape deviations and missing regions. 
The F-score, defined as the harmonic mean of precision and recall, evaluates how well generated and ground truth points lie within a threshold distance of each other.
Volume IoU voxelizes 3D models and measures their volumetric overlap.
While effective for assessing geometric accuracy, these metrics remain sensitive to outliers and errors from 3D representation transformations.
\begin{figure*}[t]
    \centering
    \includegraphics[width=\linewidth]{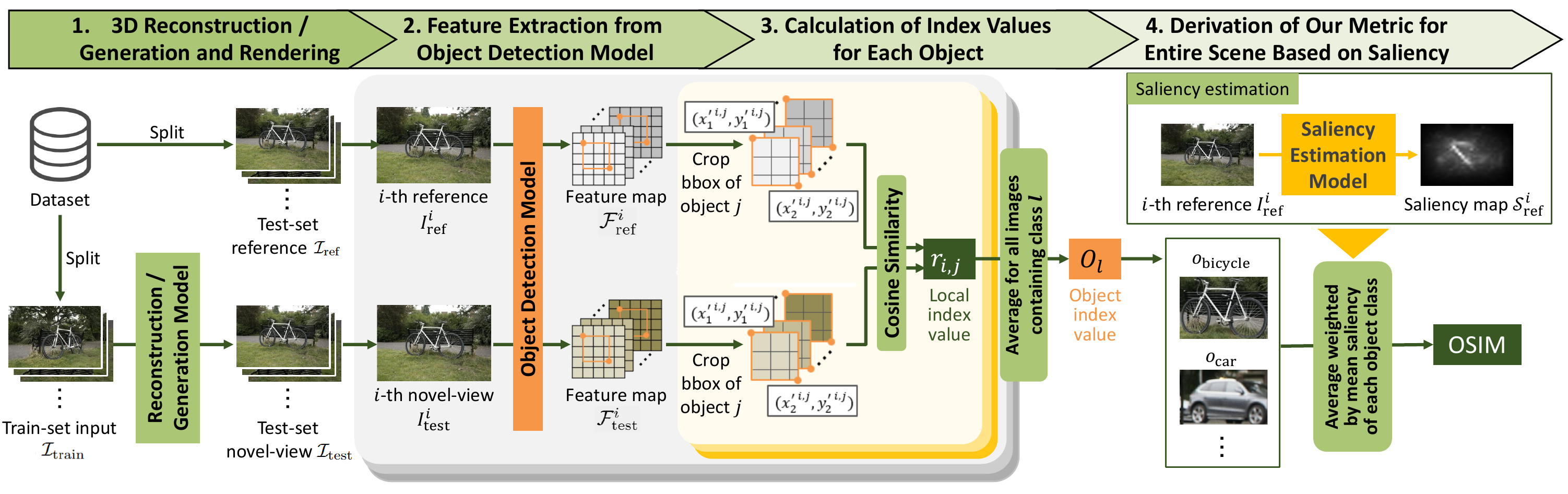}
    \caption{\small{Overview of our proposed metric. First, we split the dataset and perform reconstruction or generation followed by rendering. Second, we detect objects in the scene using an object detection model and compare the intermediate features of the detected patches between novel-view images and reference images, and obtain an index score for each object. Finally, we compute OSIM by weighting object scores according to their saliency, so that objects more likely to attract human attention contribute more to the overall score.}}
    \label{fig:overview}
\end{figure*}

\section{Objectness SIMilarity (OSIM)}
\label{sec:proposed_metric}

Figure~\ref{fig:overview} shows an overview of the proposed metric. 
OSIM consists of four steps, which are respectively described in detail in Subsecs.~\ref{sec:3.1}-\ref{sec:3.4}.
OSIM is designed to be simple and normalized. For simplicity, the metric utilizes only a pre-trained object detection model as its deep learning component.
OSIM is also designed with a normalized range of [0, 1], achieving 1.0 when the target perfectly matches the reference, following principles similar to SSIM~\cite{ssim}. 

\subsection{3D Reconstruction/Generation and Rendering}
\label{sec:3.1}
We first apply consistent pre-processing to each dataset in order to conduct a comprehensive evaluation across multiple datasets.
For reconstruction datasets, if a dataset is not pre-split, we select every $n$-th image as test data and use the remaining images for training, following the approach in~\cite{mipnerf360, 3dgs}. 
For generation datasets, we render reference images of the original 3D models at consistent intervals of $\tau$ degrees in both elevation and azimuth. 
While conventional evaluations often consider only the equatorial plane, our proposed metric also incorporates elevation angles to more accurately assess the quality of generated 3D objects.

After preparing input and reference images, both models follow a similar process for reconstruction/generation and rendering.
We input a set of $N_{\rm{train}}$ training images,  $\mathcal{I}_{\rm{train}}=\{{I}_{\rm{train}}^i \mid i=1, \ldots, N_{\rm{train}}\}$, into a 3D reconstruction or generation model to obtain a 3D scene or an object, where $i$ denotes the index of each image.
Then we render $N_{\rm{test}}$ novel-view images of the resulting scene or object from the same angles as the corresponding test set reference images, $\mathcal{I}_{\rm{ref}}= \{{I}_{\rm{ref}}^i \mid i = 1, \ldots, N_{\rm{test}}\}$, and denote these novel-view images as $\mathcal{I}_{\rm{test}} = \{{I}_{\rm{test}}^i \mid i = 1, \ldots, N_{\rm{test}}\}$. 

\subsection{Feature Extraction from Detection Model}
\label{sec:3.2}
Object detection models extract and focus on specific regions to determine object classes. 
Given that their learning and recognition processes naturally resemble human scene perception, we leverage them for our evaluation.
We use an object detection model $\mathcal{M}$ to both the test set reference image $I_{\rm{ref}}^i \in \mathcal{I}_{\rm{ref}}$ and the novel-view image $I_{\rm{test}}^i \in \mathcal{I}_{\rm{test}}$.
For each image $I_{\rm{ref}}^i$ and \( I_{\rm{test}}^i \), we extract the intermediate features from a detection model $\mathcal{M}$ and denote them as \( \mathcal{F}_{\rm{ref}}^i \in \mathbb{R}^{{W}\times{H}\times{D}} \) and \( \mathcal{F}_{\rm{test}}^i \in \mathbb{R}^{{W}\times{H}\times{D}} \), respectively. 
These features are represented as feature maps of resolution \( W \times H \), where each pixel has a feature dimension of \( D \).
Additionally, we extract the bounding box coordinates of $j$-th detected object from $i$-th reference image \( I_{\rm{ref}}^i \) based on the detection results.
These coordinates are denoted as \( ({x_1}^{i, j}, {y_1}^{i, j}) \) and \( ({x_2}^{i, j}, {y_2}^{i, j}) \), representing the upper-left and lower-right corners, respectively.
Then we calculate the corresponding positions on the feature map $ ({x_1^\prime}^{i, j}, {y_1^\prime}^{i, j})$ and $({x_2^\prime}^{i, j}, {y_2^\prime}^{i, j})$, on the basis of the resolution ratio.
Since the features of the object detection model contain essential information for object identification, OSIM can evaluate whether each object has sufficient quality to be correctly classified, which we term the ``objectness'' criterion.

\subsection{Calculation of Index Values for Each Object}
\label{sec:3.3}

Next, we calculate an index value \( r_{i, j} \) for \( j \)-th object in \( i \)-th novel-view image \( I_{\rm{test}}^i \).
We define local index value $r_{i, j}$ as the average cosine similarity of features at each pixel within the region of the feature map corresponding to the object's bounding box.
Cosine similarity is adopted to ensure the value is normalized between 0 and 1, consistent with our design principle.
The local index value $r_{i, j}$ is given by the following equation:
\begin{multline}
    r_{i, j} = \frac{1}{({x_2^\prime}^{i, j}-{x_1^\prime}^{i, j}+1)({y_2^\prime}^{i, j}-{y_1^\prime}^{i, j}+1)} \\ 
    \times \sum\limits_{x={x_1^\prime}^{i, j}}^{{x_2^\prime}^{i, j}} \sum\limits_{y={y_1^\prime}^{i, j}}^{{y_2^\prime}^{i, j}}
    cos({{{\boldsymbol{f}_{\rm{ref}}^i(x,y)}}, 
        {\boldsymbol{f}_{\rm{test}}^i(x,y)}}), 
\end{multline}
where ${\boldsymbol{f}_{\rm{ref}}^i(x, y)} \in \mathbb{R}^{D}$ and ${\boldsymbol{f}_{\rm{test}}^i(x,y)} \in \mathbb{R}^{D} $ denote the feature values of $\mathcal{F}_{\rm{ref}}^i$ and $\mathcal{F}_{\rm{test}}^i$ at position $(x, y)$, respectively, and $cos(\cdot, \cdot)$ represents the cosine similarity.

Subsequently, we calculate the index values for each object in the scene, which is one of the novelties of our metric. 
From the detection result of $\mathcal{I}_{\rm{ref}}$, we extract the classes of all detected objects and denote their union as $\mathcal{C}_{\rm{total}}$, \textit{i.e.}, the comprehensive list of all objects present in the scene.
For each object class $l \in \mathcal{C}_{\rm{total}}$, we extract index pairs \((i, j) \), where the detected object $j$ corresponds to the class $l$, and denote this set as $\mathcal{R}_l$.
The index value $o_l$ for the object class $l$ is defined as the average of $\{r_{i,j} \mid  (i, j) \in \mathcal{R}_l\}$, and is given by the following equation:
\begin{equation}
    o_l = \frac{1}{|\mathcal{R}_l|}
            \sum\limits_{(i,j) \in \mathcal{R}_l}r_{i,j}.
\end{equation}

\subsection{Derivation of Our Metric for Entire Scene Based on Saliency}
\label{sec:3.4}

Finally, we compute the overall scene-level metric.  
Since humans focus on individual objects, some naturally attract more attention than others. 
To capture this, we utilize a saliency map.
We obtain saliency map $\mathcal{S}_{\rm{ref}}^i$ for $i$-th resized reference image.
For each object $j$ in \( i \)-th reference image, we calculate its average saliency $s_{i, j}$ as follows:
\begin{multline}
    s_{i, j} = \frac{1}{({x_2}^{i, j}-{x_1}^{i, j}+1)({y_2}^{i, j}-{y_1}^{i, j}+1)}\\
    \times \sum\limits_{x={x_1}^{i, j}}^{{x_2}^{i, j}} \sum\limits_{y={y_1}^{i, j}}^{{y_2}^{i, j}}
    s_{\rm{ref}}^i(x, y),
\end{multline}
where $s_{\rm{ref}}^i(x, y)$ is the saliency at each pixel $(x, y)$ in the saliency map $\mathcal{S}_{\rm{ref}}^i$. 
Then we calculate the average saliency ${s_l}$ of each object class $l$ within the scene as follows: 
\begin{equation}
    s_l = \frac{1}{|\mathcal{R}_l|}
            \sum\limits_{(i,j) \in \mathcal{R}_l}s_{i,j}.
\end{equation}

The final metric value, OSIM, is computed as the weighted average of the index values for all classes $l$, with weights given by their average saliency.  
Therefore, OSIM is given by the following equation:
\begin{equation}
    \text{OSIM} = \frac{1}{s_{\rm{total}}}\sum\limits_{l \in \mathcal{C}_{\rm{total}}}{s_l}{o_l}, 
\end{equation}
where $s_{\rm{total}}=\sum\limits_{l \in \mathcal{C}_{\rm{total}}}{s_l}$ is the sum of the average saliency over all $l \in \mathcal{C}_{\rm{total}}$. 
Using object detection models as described above, we develop an evaluation metric focused on the ``objectness'' of a 3D scene.  
By leveraging intermediate feature representations, this metric captures object characteristics simply yet effectively, providing a more detailed axis for assessing 3D scenes.

\section{Experiments}
\subsection{Experimental Settings}

\begin{table}[t]
    \centering
    \small
    \begin{tabular}{lc|ccc}
    \hline
    \textbf{Method} & \textbf{Date} & \cite{mipnerf360} & \cite{nerf} & \cite{tanks_and_temples} \\ \hline
    COLMAP~\cite{COLMAP1, COLMAP2} & Sep. 2016 & \checkmark & \checkmark & \checkmark\\
    NeRF~\cite{nerf} & Mar. 2020 &  & \checkmark &\\
    Mip-NeRF360~\cite{mipnerf360} & Nov. 2021 & \checkmark & \checkmark &\\
    Instant NGP~\cite{INGP} & Jan. 2022 & \checkmark & \checkmark & \checkmark\\
    TensoRF~\cite{TensoRF} & Mar. 2022 &  & \checkmark\\
    NerfStudio~\cite{NeRFStudio} & Feb. 2023 & \checkmark & \checkmark & \checkmark\\
    K-Planes~\cite{K-Planes} & Jan. 2023 &  & \checkmark\\
    Zip-NeRF~\cite{Zip-NeRF} & Apr. 2023 & \checkmark & \checkmark & \checkmark\\
    Tetra-NeRF~\cite{Tetra-NeRF} & Apr. 2023 &  & \checkmark\\
    3DGS~\cite{3dgs} & Aug. 2023 & \checkmark & \checkmark & \checkmark\\
    Mip-Splatting~\cite{Mip-splatting} & Nov. 2023 & \checkmark & \checkmark & \checkmark\\
    Scaffold-GS~\cite{Scaffold-GS} & Nov. 2023 & \checkmark & \checkmark\\
    3DGS-MCMC~\cite{3DGS-MCMC} & Apr. 2024 & \checkmark & \checkmark\\
    GOF~\cite{GOF} & Apr. 2024 & \checkmark & \checkmark & \checkmark\\
    2DGS~\cite{2DGS} & May 2024 & \checkmark & & \checkmark\\
    gsplat~\cite{gsplat} & Sep. 2024 & \checkmark & \checkmark\\
    \hline
    \end{tabular}
    \caption{\label{tab:list_recon}List of models for evaluation in reconstruction. The right three columns indicate that the corresponding 3D reconstruction model was used in the evaluation with each dataset.}
\vspace{-0.2cm}
\end{table}

For 3D reconstruction models, we utilized Mip-NeRF360 dataset~\cite{mipnerf360}, Blender dataset~\cite{nerf}, and Tanks and Temples (TnT) dataset~\cite{tanks_and_temples}, following a previous study~\cite{3dgs}.
We selected a total of thirteen scenes that contain detectable objects. 
The evaluation was conducted on the group of reconstruction models listed in Table~\ref{tab:list_recon}, which are recently released and widely used.
We used the checkpoints and rendered images provided by NerfBaselines~\cite{nerfbaselines}, as they offer reconstruction results under standardized conditions.

For 3D generation models, we utilized 3D objects from the Google Scanned Objects (GSO) dataset~\cite{gso}, following previous studies~\cite{triplanegaussian, one-2-3-45++, triposr, crm, unique3d}. 
We selected a total of 30 detectable objects from this dataset, which is the same number used in~\cite{dreamgaussian, lgm, crm}. 
As test set reference images, we rendered objects at 15-degree intervals in both elevation and azimuth at a resolution of 512 $\times$ 512 pixels. 
This resulted in a total of 267 images, representing more viewpoints than any prior work to our knowledge.
We evaluated six single image-to-3D generation models: DreamGaussian (DG)~\cite{dreamgaussian}, One-2-3-45++~\cite{one-2-3-45++}, Triplane Gaussian (TGS)~\cite{triplanegaussian}, LGM~\cite{lgm}, CRM~\cite{crm}, and Unique3D~\cite{unique3d}.

As the object detection model $\mathcal{M}$, we employed YOLOX~\cite{yolox}, specifically the pre-trained YOLOX-x model with 99.1 million parameters and a mean average precision (mAP) of 51.5. 
We set the detection confidence threshold to 0.35 based on empirical results and adopted the original settings for the rest. 
The intermediate features were derived from the dark5 layer of the backbone since it represents global image context and contains information necessary for category identification. 
For saliency calculation, we used Graph-Based Visual Saliency (GBVS)~\cite{gbvs}, which is simple yet achieves high accuracy among non-deep learning methods.
In addition, GBVS is inspired by the mechanisms of human visual attention, making it well-suited for the purpose of this metric. 

We compared our proposed metric with PSNR, SSIM~\cite{ssim}, LPIPS~\cite{lpips} (AlexNet~\cite{alexnet}), and CLIP-sim. (ViT-B/32)~\cite{clip} for reconstruction models, computing each metric as the average over all novel-view test images.  
For single image-to-3D generation, we used the same metrics, supplemented by FID~\cite{fid}, Chamfer Distance (CD)~\cite{chamferdistance}, F-score~\cite{tanks_and_temples, fscore2} (threshold 0.05), and Volume IoU (V-IoU).

\subsection{Experimental Results}
\subsubsection{Exploring Evaluation Perspective of OSIM}
\begin{figure}
    \centering
    \includegraphics[width=\linewidth]{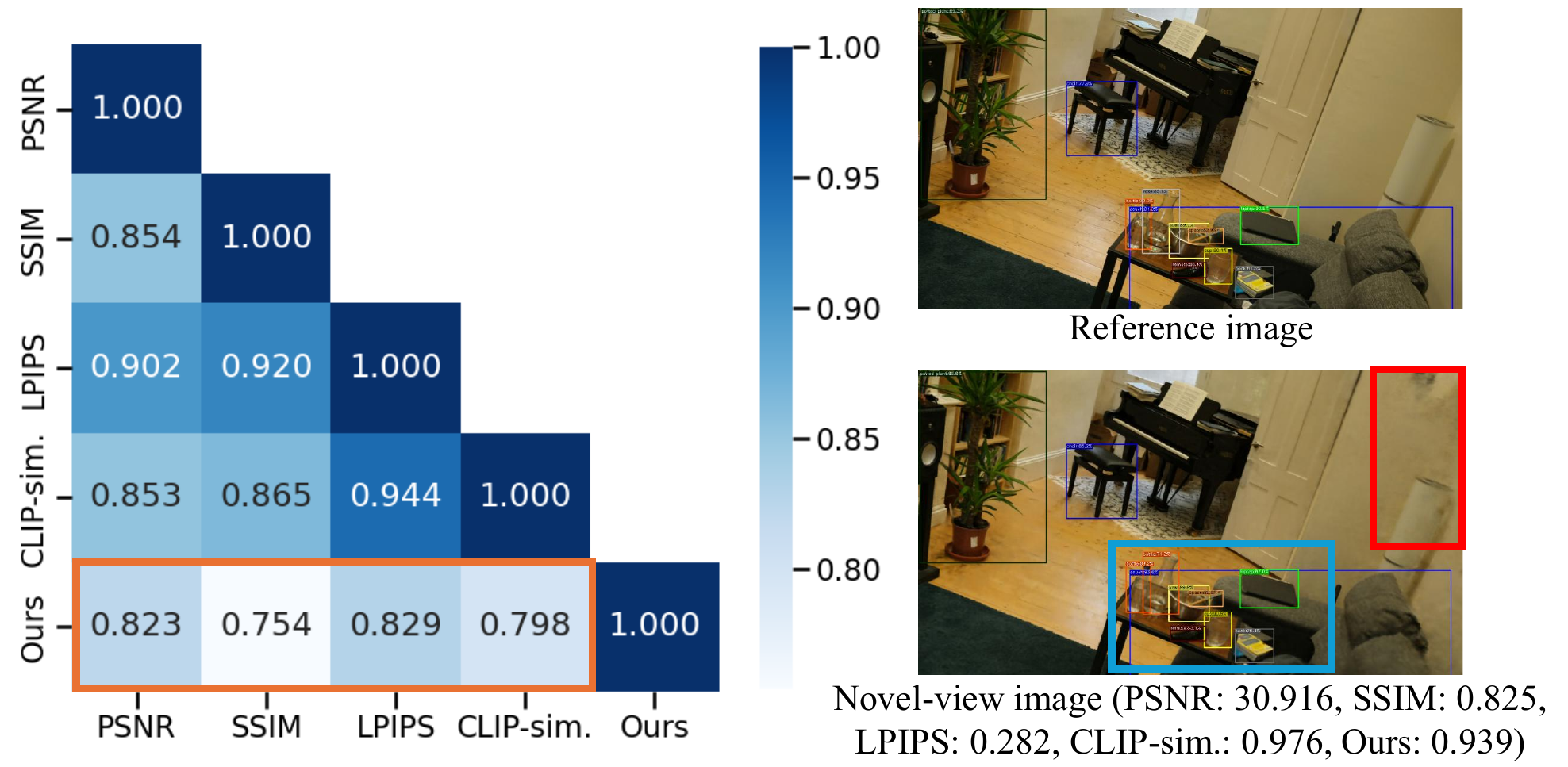}
    \caption{(left): Average Pearson correlation coefficients on the Mip-NeRF360 dataset~\cite{mipnerf360}, with the orange area indicating correlations between our proposed metric and the others. (right): A visual example highlighting OSIM's unique evaluation perspective. }
    \label{fig:corr_pearson}
\end{figure}

First, we examine whether our proposed metric evaluates from a different perspective compared to conventional metrics.
We report the average Pearson correlation coefficients~\cite{pearson_corr} for each pair of metrics.
For metrics where lower values indicate better performance, the correlation was computed after multiplying the values by $-1$.
In addition, we excluded COLMAP~\cite{COLMAP1} from the correlation calculation, as this older model, proposed in 2016, was treated as an outlier due to abnormally low values across all metrics.
As shown on the left of Fig.~\ref{fig:corr_pearson}, the correlation between conventional metrics and our proposed metric OSIM is lower than those among conventional metrics themselves, indicating that OSIM evaluates from a distinct perspective.

The right side of Fig.~\ref{fig:corr_pearson} shows a visual example illustrating how OSIM evaluates from a different perspective.
In the novel-view image, the red box highlights wall smudging, which lowers scores in conventional visual quality metrics.
In contrast, the blue box shows objects on the table reconstructed with sufficient quality for detection, resulting in a high OSIM value.
Since OSIM does not penalize background noise, it assigns high scores as long as objects are reconstructed with high quality.
This example underscores the importance of employing multiple metrics to capture different evaluation perspectives.
When combined with conventional metrics, OSIM enables both object-focused evaluation and a more comprehensive assessment that also accounts for the background.

\begin{figure}[t]
    \centering
    \includegraphics[width=\linewidth]{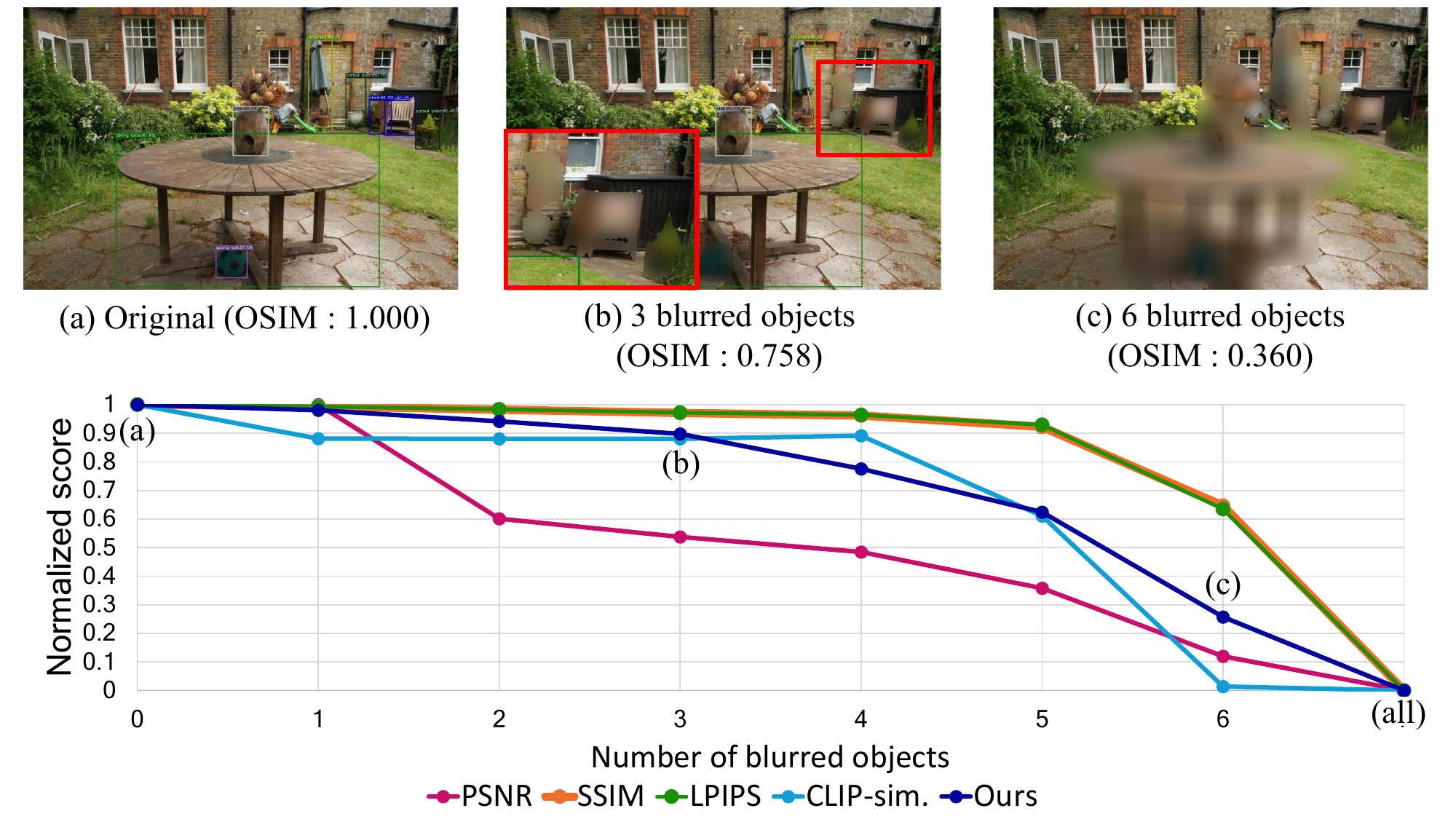}
    \caption{\small{Evaluation results with added Gaussian blur. From (a) to (c), more blur leads to fewer detections and lower metric values.}}
    \label{fig:variation_with_noise}
\end{figure}

\subsubsection{Metric Variation with Degradations}
We evaluate the sensitivity of the proposed metric with respect to image degradations.
Specifically, in scenes containing multiple objects, Gaussian blur is applied incrementally to each object, from smaller to larger sizes, gradually degrading image quality.
This procedure allows us to rigorously assess how the proposed metric responds to varying levels of degradation.
Additionally, we normalize the metric values for each image to the range [0, 1], scaling them between the highest score and the score obtained when degradations are applied to the entire image, including the background.
This normalization facilitates comparison of score decreases across metrics with different value ranges (\textit{e.g.}, PSNR has a distinct range, whereas CLIP-sim. tends to yield higher scores).
For PSNR, calculation began with one blurred object, since the metric value for the reference image becomes infinite.
For LPIPS, the final value was obtained by subtracting the score from 1.
Figure~\ref{fig:variation_with_noise} shows how metric values change as the number of blurred objects increases.
OSIM decreases consistently with blur, while PSNR also decreases but shows large drops for objects 1 and 2.
LPIPS and SSIM follow similar trends, with sharp drops when blur affects large objects.
In contrast, OSIM exhibits strong drops in cases like (b) and (c), where salient objects (\textit{e.g.}, the vase on the table) are blurred, showing that it captures degradations overlooked by conventional metrics.
CLIP-sim., however, sometimes increases or remains nearly unchanged until four objects are blurred, making it unreliable for assessing degradation.

\subsubsection{Effect of Object-Level Evaluation}
\label{subsubsec:object-level}

\begin{figure}[t]
    \centering
    \includegraphics[width=\linewidth]{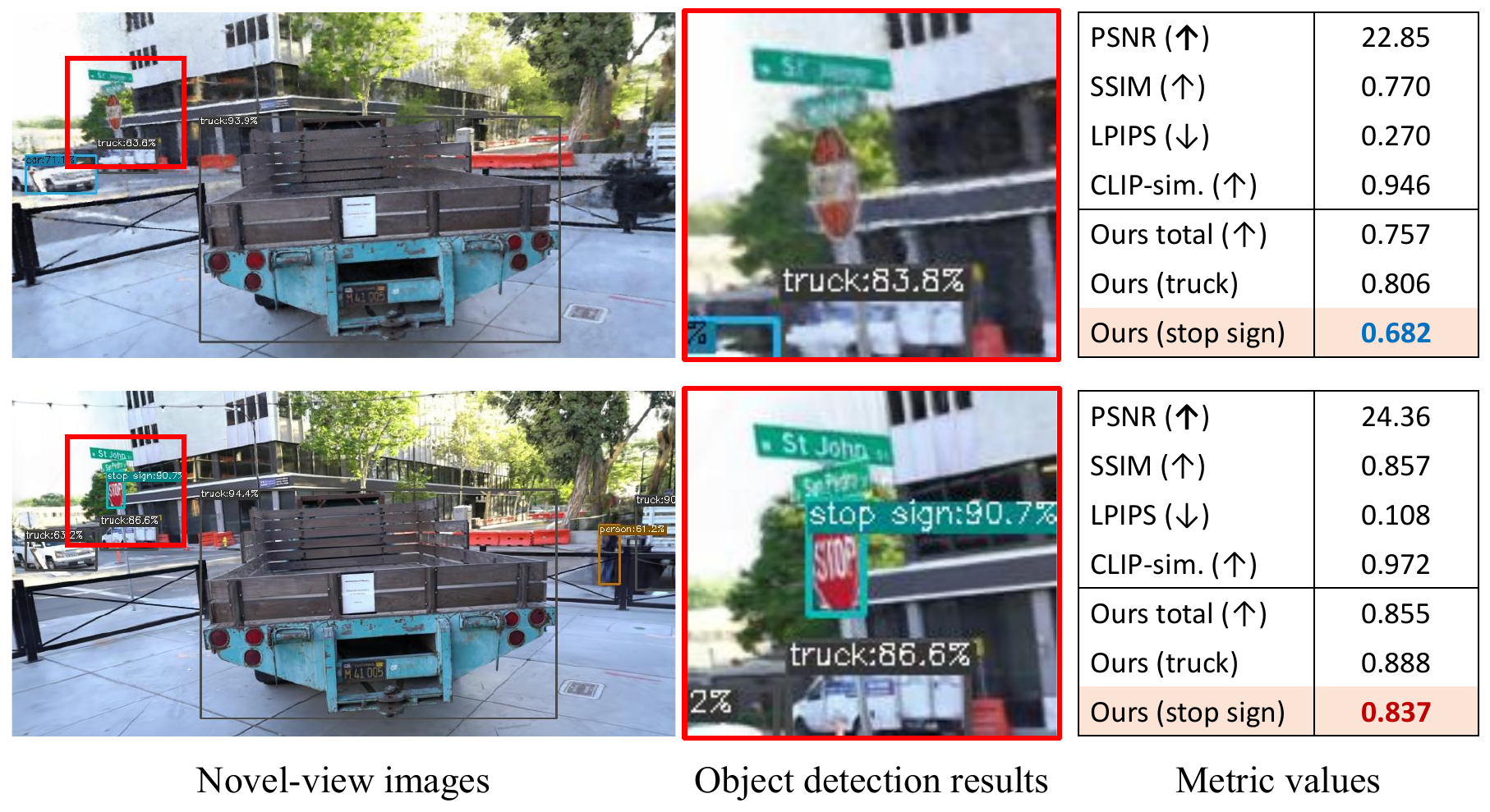}
    \caption{\small{Object-level evaluation examples. (Top): Most objects are accurately reconstructed, but the stop sign shows low quality and is not detected. (Bottom): All objects are well reconstructed.}}
    \label{fig:real_data}
\end{figure}

\begin{figure}[t]
    \centering
    \includegraphics[width=\linewidth]{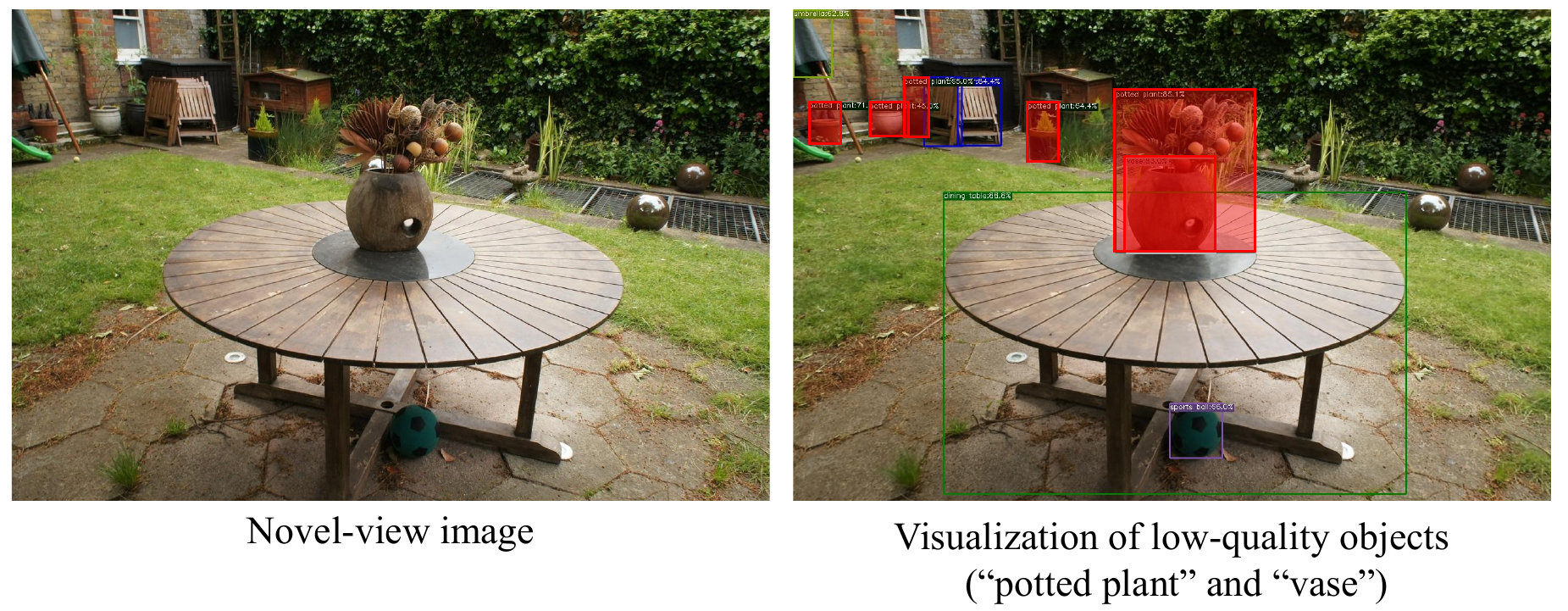}
    \caption{Visualization of low-quality objects with semi-transparent red masks indicating those scoring below the scene-level OSIM value.}
    \label{fig:low-quality}
\end{figure}

The top of Fig.~\ref{fig:real_data} shows a scene containing both high- and low-quality objects.
Conventional metrics assess only overall quality, often yielding intermediate values that obscure which objects are of low quality.
In contrast, the proposed metric evaluates each object individually, assigning high scores to high-quality objects and low scores to low-quality ones.
In the top example, the overall reconstruction appears successful, but the ``stop sign'' lacks sufficient quality to be accurately predicted, with OSIM (stop sign) = 0.682. 
In the bottom example, all objects, including the ``stop sign,'' are well reconstructed, resulting in a higher OSIM (stop sign) score of 0.837.
This results demonstrate that the proposed object-level metric effectively assesses the quality of individual objects in complex scenes.
Although the scene-level OSIM falls in an intermediate range, similar to conventional metrics, this aligns with human perception.
The key advantage of OSIM lies in its ability to evaluate objects individually, enabling a more detailed quality assessment.

The proposed metric also facilitates intuitive and semantically meaningful visualization.
With conventional metrics, visualizing regions below a quality threshold often produces scattered or fragmented pixels, making interpretation difficult. 
Deep model–based metrics are even harder to visualize meaningfully.
In contrast, as shown in Fig.~\ref{fig:low-quality}, our metric overlays bounding box masks on low-quality objects, allowing users to clearly identify which objects require improvement in reconstruction or generation quality.

\subsection{User Study and Analysis}
\label{sec:user_study}

\begin{table*}[t]
    \centering
    \small
    \begin{tabular}{l|cccc|ccccc}
         \hline
         \small
         Calculation method & \multicolumn{4}{c|}{Whole image} & \multicolumn{5}{c}{Bbox Patch} \\
         \hline
         Evaluation metrics & PSNR & SSIM & LPIPS & CLIP-sim. & PSNR & SSIM & LPIPS & CLIP-sim. & Ours\\
         \hline
         $\rho$ & 0.658 & 0.483 & 0.647 & 0.774 & 0.579 & 0.551 & 0.702 & 0.756 & \textbf{0.820}\\
         \hline
    \end{tabular}
    \caption{\label{tab:corr_mip}\small{Average Spearman's rank correlation coefficients ($\rho$)~\cite{spearman_rank} between the MOS and each metric in reconstruction models. }}
\end{table*}

\begin{table*}[t]
    \centering
    \small
    \begin{tabular}{l|cccccc|ccc}
         \hline
         \small
         Metric Type & \multicolumn{6}{c|}{Image-based} & \multicolumn{3}{c}{3D Shape-based}\\
         \hline
         Evaluation metrics & PSNR & SSIM & LPIPS & CLIP-sim. & FID & Ours & CD & V-IoU & F-score \\
         \hline
         $\rho$ & 0.314 & 0.600 & 0.600 & 0.714 & 0.714 & \textbf{0.943} & 0.771 & 0.600 & 0.711 \\
         \hline
    \end{tabular}
    \caption{\label{tab:corr_gso}\small{Average Spearman's rank correlation coefficients ($\rho$)~\cite{spearman_rank} between the MOS and each metric in generation models.}}
\end{table*}

Since 3D reconstruction and generation models aim to produce outputs perceived as high-quality by humans, evaluation metrics must align with human perception.
To verify this, we conducted a user study with 23 participants from our research lab. 
For reconstruction models, we used seven scenes from the Mip-NeRF360 dataset~\cite{mipnerf360} and 12 models (Table~\ref{tab:list_recon}), yielding 84 reconstructed scenes.
In each trial, rendered images of a randomly selected scene were shown with reference images, and participants rated them on a five-point scale to obtain Mean Opinion Scores (MOS), considering visual quality, objectness, and semantic fidelity.
Each participant completed ten trials, resulting in a total of 230 evaluations. 
Unlike prior studies~\cite{mipnerf360, evaluatingNVS}, which used 360-degree or front-facing videos, we presented novel-view and reference images as slideshows to unify study conditions with objective metrics.
For generation models, we selected 30 objects from the GSO dataset~\cite{gso} and generated them with six single-image-to-3D models, resulting in 180 samples.
In each trial, a generated object was shown as a 360-degree rotation video along the equatorial plane, accompanied by its input image.
Participants rated each object on the same five-point MOS scale, considering visual quality, objectness, and structural fidelity.
Each participant completed 20 trials, resulting in a total of 460 evaluations.
Evaluation was based on the following criteria:
\begin{itemize}
  \item \textbf{Visual Quality}: This criterion assesses overall visual quality. For reconstruction models: consider edge blurriness, haze artifacts, floor patterns, and reflections. For generation models: evaluate both visible regions and unseen perspectives that are not present in the input image.
  \item \textbf{Objectness}: This creiterion assesses whether each object is recognizable and identifiable as its class. (Ignore background quality; focus only on objects.)
  \item \textbf{Semantic Quality}: This criterion evaluates how accurately the semantics of the reference/input image is reproduced. (Assess both the objects and the background.)
  \item \textbf{Structural Quality}: This criterion measures the quality of the 3D shape. A lower scores should be given for extra parts, disconnected fragments, or rough surfaces.
  \item \textbf{Overall Quality}: This criterion evaluates the overall quality and suitability for real-world applications.
\end{itemize}

We report the Spearman's rank correlation coefficient ($\rho$)~\cite{spearman_rank} between the MOS and each metric score. 
Results for reconstruction models are shown in Table~\ref{tab:corr_mip}.
To fairly compare with our proposed metric, which is computed for each object-detected patch, conventional metrics were also evaluated in a patch-based manner.
As shown in Table~\ref{tab:re-eval_mip}, since COLMAP scores were extremely low and treated as outliers, we excluded them from the calculation.
OSIM achieves the highest rank correlation among all metrics, indicating it better captures human perceptual preferences and supporting its validity as a reliable evaluation metric.
Moreover, we found that conventional metrics, except for PSNR, also exhibit higher correlations with human judgments when evaluated on object patches rather than whole images.
This confirms the importance of object-focused evaluation, while the superior performance of OSIM further demonstrates the effectiveness of using features from object detection models.
Results for generation models are presented in Table~\ref{tab:corr_gso}.
Since participants assessed the models by watching 360-degree rotating videos, we computed each metric using only the equatorial rendered images to match this setting.
OSIM again achieves the highest rank correlation, aligning closely with human perception.
In addition to deep model–based metrics such as CLIP-sim. and FID, shape-based metrics like CD and F-score also show high correlations, suggesting that considering 3D geometry is crucial for the evaluation of generative models.

\subsection{Re-evaluation of 3D Reconstruction Models}
\begin{table*}[t]
    \centering
    \small
    \begin{tabular}{lc|cccccccc}
    \hline	
	Method|Metric
        &Date& PSNR$\uparrow$ & SSIM$\uparrow$ & LPIPS$\downarrow$ & CLIP-sim.$\uparrow$ & Ours$\uparrow$ & MOS$\uparrow$ & Train$\downarrow$ & Mem $\downarrow$ \\
	\hline\hline 
        COLMAP~\cite{COLMAP1, COLMAP2} & Sep. 2016 & 16.556 & 0.483 & 0.576 & 0.802 & 0.548 & 1.710 & 2h59m42s & \cellcolor{red!40}0.00GB \\
        Mip-NeRF360~\cite{mipnerf360}&Nov 2021 & 28.700 & 0.828 & 0.254 & \cellcolor{yellow!40}0.984 & \cellcolor{orange!40}0.950 & \cellcolor{red!40}4.500 & 30h14m37s & 33.83GB \\
        INGP~\cite{INGP}& Jan. 2022 & 26.454 & 0.729 & 0.381 & 0.965 & 0.903 & 3.500 & \cellcolor{red!40}3m56s & 8.25GB \\
        NerfStudio~\cite{NeRFStudio}&Feb. 2023 & 27.219 & 0.767 & 0.328 & 0.973 & 0.930 & 3.667 & \cellcolor{orange!40}19m38s & \cellcolor{orange!40}6.05GB \\
        Zip-NeRF~\cite{Zip-NeRF}&Apr. 2023 & \cellcolor{red!40}29.606 & \cellcolor{red!40}0.861 & \cellcolor{red!40}0.203 & \cellcolor{red!40}0.988 & \cellcolor{red!40}0.962 & \cellcolor{orange!40}4.400 & 5h33m2s & 26.85GB \\
        3DGS~\cite{3dgs}&Aug. 2023 & 28.389 & 0.850 & 0.242 & \cellcolor{yellow!40}0.984 & 0.941 & 4.333 & \cellcolor{yellow!40}23m54s & 11.10GB \\
        Mip-Splatting~\cite{Mip-splatting}&Nov. 2023 & 28.451 & 0.851 & 0.243 & \cellcolor{yellow!40}0.984 & 0.940 & 4.000 & 26m5s & 10.95GB  \\
        Scaffold-GS~\cite{Scaffold-GS}&Nov. 2023 & \cellcolor{orange!40}28.769 & \cellcolor{yellow!40}0.852 & 0.243 & 0.983 & 0.939 & 4.294 & 24m50s & 9.33GB \\
        3DGS-MCMC~\cite{3DGS-MCMC}&Apr. 2024 & \cellcolor{orange!40}28.769 & 0.850 & 0.251 & 0.975 & 0.911 & 4.167 & 37m51s & 22.04GB \\
        GOF~\cite{GOF}&Apr. 2024 & 28.300 & \cellcolor{orange!40}0.856 & \cellcolor{orange!40}0.225 & \cellcolor{orange!40}0.985 & \cellcolor{yellow!40}0.945 & 4.182 & 1h7m3s & 26.78GB \\
        2DGS~\cite{2DGS}&May 2024 & 27.714 & 0.834 & 0.281 & 0.979 & 0.925 & 3.955 & 32m6s & 11.75GB \\
        gsplat~\cite{gsplat}&Sep. 2024 & 28.364 & 0.851 & \cellcolor{yellow!40}0.241 & \cellcolor{yellow!40}0.984 & 0.940 & \cellcolor{orange!40}4.400 & 30m9s & \cellcolor{yellow!40}8.06GB \\ \hline
    \end{tabular}
    \caption{\label{tab:re-eval_mip}\small{Re-evaluation results of 3D reconstruction models on Mip-NeRF360 dataset~\cite{mipnerf360}. The score of conventional metrics, train time (Train) and GPU memory (Mem) are from NerfBaselines~\cite{nerfbaselines}.}}
\end{table*}

\begin{figure*}[t]
    \centering
    \begin{minipage}{.55\textwidth}
        \includegraphics[width=\linewidth]{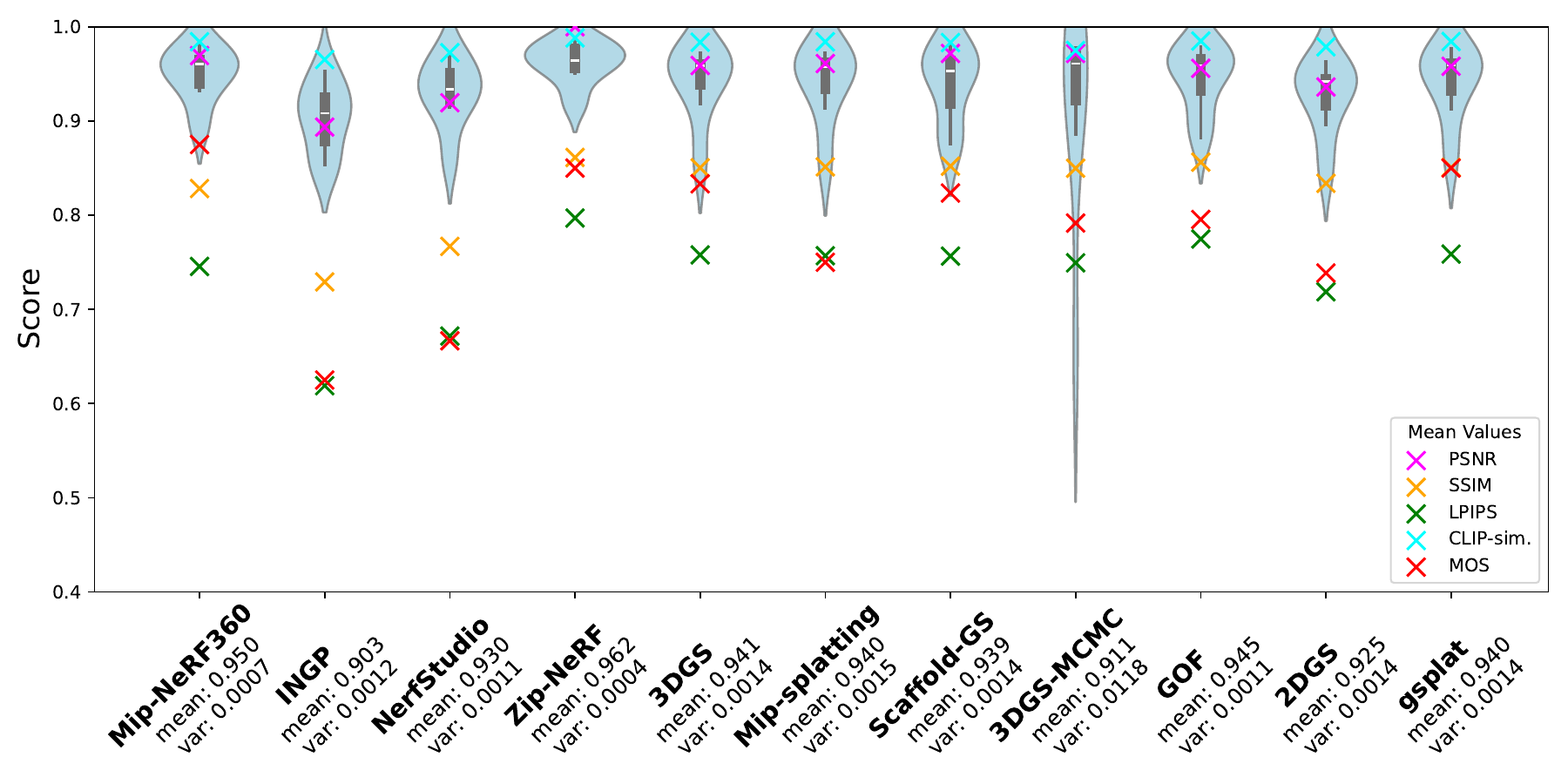}
        \subcaption{Reconstruction models on Mip-NeRF360 dataset~\cite{mipnerf360}}
    \end{minipage}%
    \begin{minipage}{.45\textwidth}
        \includegraphics[width=\linewidth]{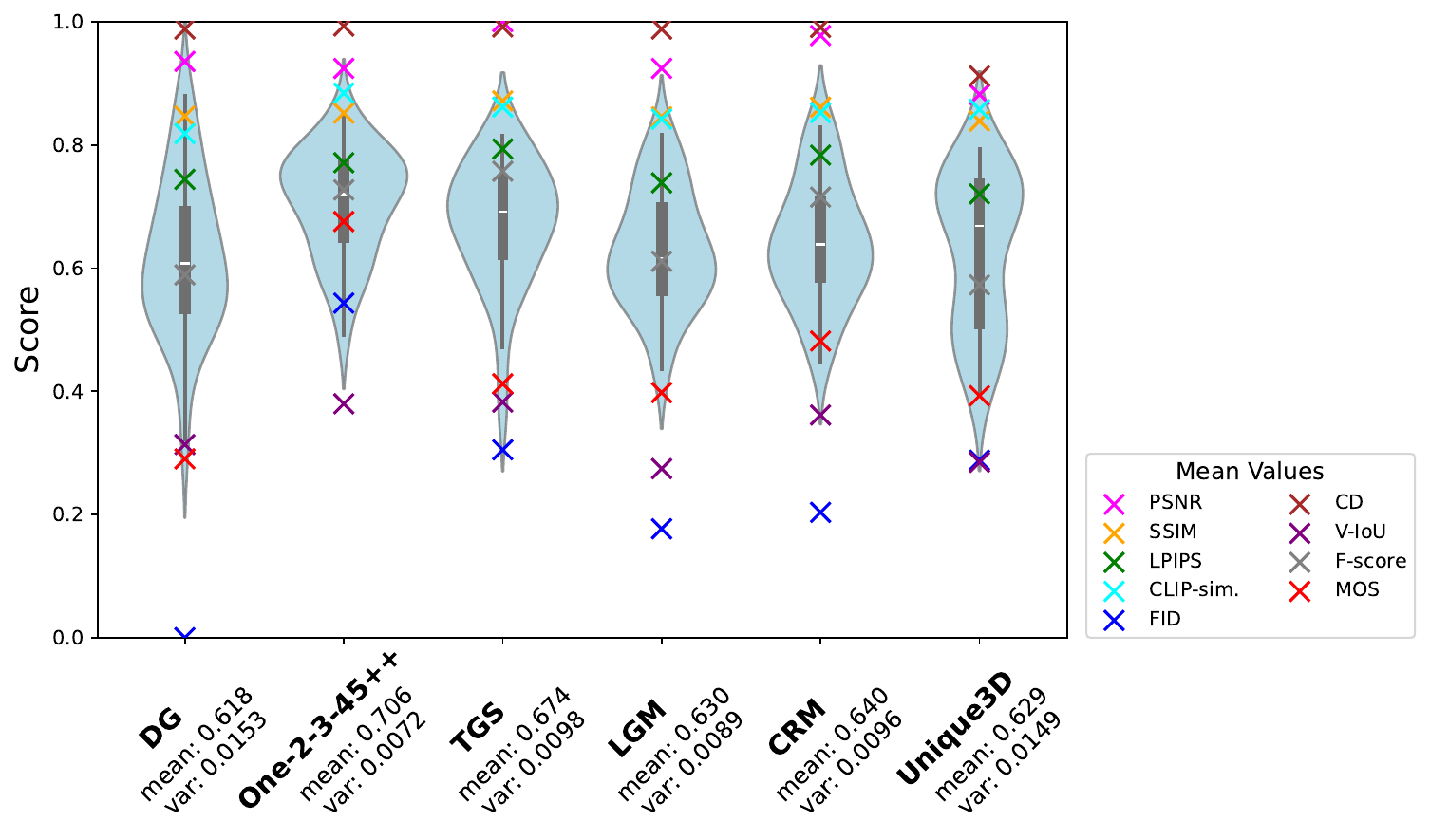}
        \subcaption{Generation models on GSO dataset~\cite{gso}}
    \end{minipage}%
    \caption{Re-evaluation results on Mip-NeRF360 dataset and GSO dataset on our proposed metric. The models are arranged from left to right in order of the publication dates. The violin plots show the box plot, variance, and density of each model's OSIM values. The symbol ``×" indicates the normalized average value of the comparative metrics.}
    \label{fig:re-eval-mip_6}
\end{figure*}

We show the comprehensive re-evaluation results in Table~\ref{tab:re-eval_mip} and Fig.~\ref{fig:re-eval-mip_6}. 
From Table~\ref{tab:re-eval_mip}, Zip-NeRF~\cite{Zip-NeRF} is highly rated across all metrics including the MOS, indicating its high quality.
Mip-NeRF360~\cite{mipnerf360}, which has the highest MOS, is highly rated only by the proposed metric and CLIP-sim., suggesting a strong correlation between the proposed metric and the MOS. 
However, both Zip-NeRF and Mip-NeRF360 require substantial training time and GPU memory. 
Considering this trade-off, gsplat and 3DGS may be a reasonable choices for some applications. 

In Fig.~\ref{fig:re-eval-mip_6}, the models are sorted from left to right in order of their release dates. COLMAP~\cite{COLMAP1, COLMAP2} is excluded since its scores are extremely low and considered to be outliers. 
To normalize each metric, PSNR scores were divided by the highest score among all models, while LPIPS and CD values were subtracted from 1. 
For FID, both operations (division and subtraction from 1) were applied.
MOS values were normalized by subtracting 1 and dividing by 4.
The re-evaluation results using a comprehensive test set based on our metric reveal an interesting trend: excluding time-consuming methods such as Mip-NeRF360~\cite{mipnerf360} and Zip-NeRF~\cite{Zip-NeRF}, the emergence of 3DGS-based approaches has led to performance improvements over NeRF-based methods. 
However, among the 3DGS-based methods developed between 2023 and 2024, no dramatic improvements are observed.
This trend is also observed in generation models.
These findings show that evaluating the performance of 3D reconstruction models should not be limited to discussing numerical improvements in conventional metrics. 
Instead, a multifaceted evaluation is needed, considering both perceived quality from a human perspective and practical factors such as computational cost and surface smoothness. 
This challenge extends beyond 3D reconstruction and generation to the broader field of computer graphics and vision.

\section{Discussion}
\label{sec:discussion}

\begin{table}[t]
    \small
    \centering
    \begin{tabular}{l|cccccc}
        \hline
        layers &  dark3 & dark4 & dark5 & p3 & p4 & p5\\
	\hline 
	Recon. & 0.651 & 0.743 & \textbf{0.820} & 0.392 & 0.711 & 0.797\\
	Gen. & 0.657 & 0.657 & \textbf{0.943} & 0.429 & 0.771 & 0.771\\
        \hline
    \end{tabular}
    \caption{\small{Average Spearman's rank correlation coefficients ($\rho$)~\cite{spearman_rank} for different intermediate layers of the object detection model on the Mip-NeRF360~\cite{mipnerf360} and GSO dataset~\cite{gso}. Recon. stands for 3D reconstruction models, and Gen. stands for 3D generation models.}}
    \label{diff_layer1}
\end{table}

\noindent\textbf{Ablation Study.}
\label{sec:ablation}
\indent We present the results of OSIM’s rank correlations when using different intermediate layers of the object detection model in Table \ref{diff_layer1}.
Specifically, we employed six layers, including the backbone layers (dark3, dark4, dark5), as well as the neck layers (p3, p4, p5) arranged in order of increasing depth.
The results show that the highest rank correlation is achieved with dark5, while p5 also tends to yield relatively better performance.

\noindent\textbf{Limitations.}
\label{sec:limitations}
\indent OSIM has limitations due to its reliance on an object detection model.
For example, objects belonging to classes that the object detection model has not been trained on are not evaluated.
However, since feature maps can still be extracted from regions where objects are not detected, users can manually specify regions of interest to evaluate such objects.
In our experiments, the detection models were fairly robust to overlapping objects.
Nevertheless, distinguishing multiple instances of the same class within a single scene remains challenging.
Note that these are common limitations shared by many metrics that depend on deep learning models~\cite{lpips, clip, fid}.
In such cases, pixel-based and structure-based metrics can complement the evaluation, improving overall accuracy. 
The key contribution of OSIM lies in introducing a novel, object-centric evaluation perspective bridging the gaps left by conventional metrics. 
While introducing new evaluation perspectives is a significant contribution to the field, the reliability of the evaluation models themselves must be continually scrutinized.

\noindent\textbf{Future Work.}
\indent Our evaluation metric is considered applicable not only to 3D scenes but also to the quality assessment of 2D images. 
To explore its broader applicability, its effectiveness should be validated using datasets specifically designed for 2D image quality evaluation. 
Additionally, this metric has the potential to extend to the evaluation of 4D dynamic scenes or generated videos, enabling the object-centric assessment in time-evolving visual data. 
Future work should investigate the feasibility of such extensions and explore ways to further refine the metric for diverse visual contents.
Also, as we mentioned in Subsubsec.~\ref{subsubsec:object-level}, our proposed metric can also serve as a diagnostic tool for visualizing low-quality objects.
A promising direction is to leverage it for identifying low-quality regions at the object level and improving reconstruction or generation quality through iterative refinement of those objects.
\section{Conclusion}
We have proposed the Objectness SIMilarity (OSIM) metric for 3D scene evaluation. 
By explicitly focusing on the quality of individual objects within 3D scenes, OSIM addresses a critical limitation in existing evaluation metrics, aligning evaluations more closely with human perception. User studies show OSIM achieves the highest correlation with human subjective assessments. 
Our metric is designed to support the advancement and societal implementation of 3D reconstruction and generation technologies by improving evaluation accuracy in practical applications. 
Establishing appropriate evaluation metrics from diverse perspectives is essential not only for promoting methodological development but also for guiding the broader adoption of these technologies.

\clearpage
\noindent\textbf{Acknowledgements.}\\
This work was partly supported by the Japan Society for the Promotion of Science (JSPS) KAKENHI Grant Numbers JP24K02942, JP23K21676, JP23K11211 and JP23K11141. 
{
    \small
    \bibliographystyle{ieeenat_fullname}
    \bibliography{main}
}


\end{document}